%% file: main-arxiv.tex
\documentclass[letterpaper]{article}
\usepackage{aaai2027}
\nocopyright
\usepackage[hyphens]{url}
\usepackage{graphicx}
\usepackage{natbib}
\usepackage{caption}
\usepackage{amsmath,amssymb,amsfonts,amsthm,mathtools}
\usepackage{booktabs}
\usepackage{xcolor}
\usepackage{multirow}
\usepackage{makecell}
\usepackage{algorithmic}
\usepackage{algorithm}
\usepackage{tikz}
\usetikzlibrary{positioning}
\usepackage{array}

\theoremstyle{plain}

\theoremstyle{definition}

\theoremstyle{remark}

\input{math_commands}

\title{WM-R1: Training GUI Agents to Reason and leverage World Models with Reinforcement Learning}

\author{
    Yu Han$^1$, Tianwen Qian$^{\dagger1}$\\
    $^1$\normalfont School of Computer Science and Technology, East China Normal University \\
    \small\normalfont{10245102432@stu.ecnu.edu.cn, twqian@cs.ecnu.edu.cn}
    \thanks{$^\dagger$Corresponding author.}
}
\affiliations{}
\makeatletter
\def\thanks#1{\protected@xdef\@thanks{\@thanks
        \protect\footnotetext{#1}}}
\makeatother

\begin{document}

\maketitle

\begin{abstract}
\input{sections/0_abstract}
\end{abstract}

\input{sections/1_introduction}
\input{sections/2_related_work}
\input{sections/3_preliminaries}
\input{sections/4_method}
\input{sections/5_experiments}

\input{sections/6_conclusion}

% Acknowledgments (uncomment for camera-ready)
% \section*{Acknowledgments}
% Acknowledgments to be added upon acceptance.

\bibliography{references}

\appendix
\input{sections/A_appendix}

\end{document}

%% file: math_commands.tex
\newcommand{\EE}{\mathbb{E}}

\DeclareMathOperator{\KL}{KL}

\newcommand{\gA}{\mathcal{A}}  % action space
\newcommand{\gT}{\mathcal{T}}  % transition dynamics
\newcommand{\act}{a}          % action
\newcommand{\rew}{r}          % reward
\newcommand{\traj}{\tau}      % trajectory
\newcommand{\pol}{\pi}        % policy
\newcommand{\wm}{\mathcal{W}} % world model
\newcommand{\vphi}{\varphi}   % visual observation (screenshot)
\newcommand{\instr}{I}        % instruction
\newcommand{\hist}{h}         % interaction history

%% file: sections/0_abstract.tex
GUI agents trained with reinforcement learning (RL) have showcased strong environment learning capabilities on mobile platforms. However, RL typically demands extensive real-environment interactions, leading to high resource costs and instability, especially in GUI scenarios.
To address these, we propose WM-R1, the first reinforcement learning framework that trains mobile GUI agents with world models instead of real environments. Specifically, world models serve as the source of state transitions during all rollouts, replacing the real Android environment within the training loop. WM-R1 also embeds world models directly into the thinking process, enabling agents to reason about the consequences of candidate actions before committing to the final action.
Crucially, WM-R1 eliminates the need for real-environment interaction, supports massively parallelized and step-level granularized trajectory generation grounded in world models, and introduces a multi-dimensional rule-based reward that jointly optimizes task success, trajectory efficiency, and world model utilization.
For efficient training, we curate a high-quality dataset of 2000 challenging tasks.
Experiments on Android mobile benchmarks demonstrate that WM-R1-trained agents significantly outperform GRPO-only baselines and inference-time simulation methods.
Code is available at \url{https://github.com/genalyu/WM-R1}.

%% file: sections/1_introduction.tex
\section{Introduction}
\label{sec:introduction}

Autonomous GUI agents that interact with mobile interfaces through visual perception and action execution have emerged as a promising direction for automating complex digital workflows~\cite{qin2025uitars,cheng2024cogagent,rawles2024androidinthewild,zhang2024guisurvey3}. Reinforcement learning (RL) has proven effective for training these agents. While PPO~\cite{schulman2017ppo} requires training a separate value network, GRPO~\cite{shao2024deepseekmath} eliminates this overhead by computing advantages relative to a group of sampled trajectories, enabling emergent reasoning through rule-based rewards alone. UI-R1~\cite{lu2025uir1} was the first to apply GRPO to GUI action prediction, using a rule-based reward (format + action type + coordinate) without supervised reasoning traces. ARPO~\cite{lu2025arpo} introduced an experience replay buffer into GRPO, reusing successful trajectories for long-horizon tasks. Recent works have extended this direction with multi-turn RL~\cite{zhang2025tars2}, curriculum training~\cite{wang2025craft} and self-evolutionary strategies~\cite{li2025segui,zhang2025uiagile}.

\begin{figure}[t]
\centering
\includegraphics[width=\linewidth]{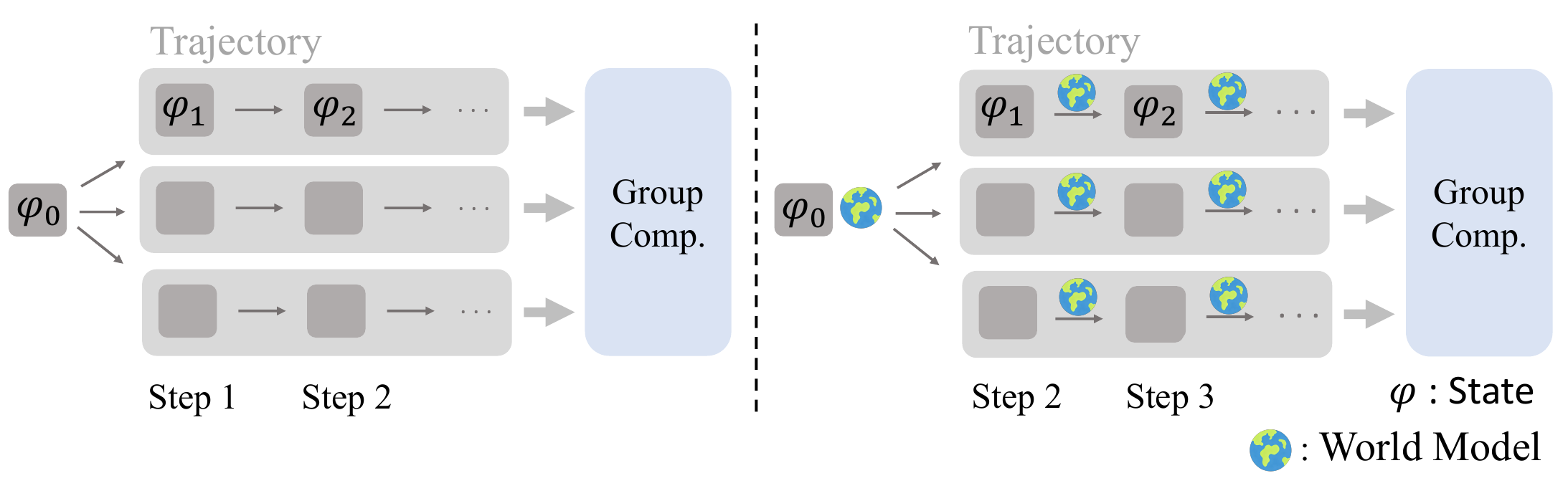}
\caption{Comparison of training paradigms. Left: Standard GRPO generates trajectories through real-environment interaction, incurring high computational cost, environment latency, and irreversibility. Right: WM-R1 replaces the environment with world models, enabling fully simulated trajectory generation.}
\label{fig:comparison}
\end{figure}

However, all existing RL methods require extensive real-environment interaction, introducing three bottlenecks: (1) \textit{computational cost}: environment interaction is orders of magnitude slower than model inference; (2) \textit{irreversibility}: a single wrong action derails long-horizon trajectories; (3) \textit{environment noise}: non-deterministic transitions corrupt the reward signal.
Parallel work has explored world models as a remedy. Text-based methods predict transitions via natural language~\cite{chae2024guiworldmodel} or DOM updates~\cite{gu2024llmwebworldmodel}, capturing semantic intent but discarding spatial-visual details. Pixel-based approaches like ViMo~\cite{luo2025vimo} and MobileDreamer~\cite{chen2026mobiledreamer} synthesize screenshots but struggle with structural controllability. Code-based methods such as Code2World~\cite{zheng2026code2world} generate renderable HTML, achieving both visual fidelity and structural precision. However, all these works use world models exclusively at inference time as a plug-and-play simulator that augments a pre-trained agent's decision-making, leaving the RL training loop unchanged. The fundamental challenge remains: how to leverage the world model not just for inference, but as the training environment itself.

To address these challenges, we propose WM-R1, the first reinforcement learning framework that trains mobile GUI agents entirely within a simulated environment, namely world models. As shown in Figure~\ref{fig:comparison} (right), WM-R1 replaces the real Android environment with world models that predict next GUI states via simulated screenshots. This substitution has two critical consequences: (1) training becomes massively parallelizable and step-level granularized with environmental interaction overhead, and (2) the agent's chain-of-thought reasoning naturally incorporates world model predictions, since every state transition it observes comes from the world model. Specifically, WM-R1 embeds the world model directly into the agent's reasoning stream through a \texttt{<call\_wm>} mechanism, allowing the agent to simulate and refine candidate actions before committing to final outputs.

Our approach uses Qwen2.5-VL-3/7B~\cite{bai2025qwen25vl} as the base model and trains on a curated dataset combining three Android GUI sources: AndroidCode~\cite{zheng2026code2world}, GUI-Odyssey~\cite{lu2024guiodyssey}, and GUI-R1~\cite{gui_r1_2025}. We filter by task difficulty. The training loop follows GRPO~\cite{shao2024deepseekmath} with a composite reward (success + length + WM call count) to drive the policy update.
Our contributions are:
\begin{enumerate}
    \item WM-R1 framework: A GRPO-based training framework that replaces the real Android environment with a learned world model, enabling fully simulated, massively parallel training of mobile GUI reasoning agents without any real-environment interaction.
    \item World model utilization: We demonstrate that WM-R1 substantially improves GUI agents' ability to reason with world models on mobile platforms.
    \item Multi-dimensional reward design: We propose a rule-based composite reward that aligns with GUI task objectives across task success, trajectory efficiency, and world model utilization, with ablation studies confirming each component's contribution.
    \item Experiments on mobile GUI benchmarks: We train on the curated Android dataset and evaluate on AndroidWorld, GUI-Odyssey, and AndroidControl, demonstrating that WM-R1-trained agents outperform GRPO baselines at both 3B and 7B model scales.
\end{enumerate}

%% file: sections/2_related_work.tex
\section{Related Work}
\label{sec:related_work}

\paragraph{GUI agents.}
Mobile GUI agents perform multi-step interactions on Android devices from visual input~\cite{zhang2024guisurvey3,liu2024guisurvey2,wang2024guisurvey}. The prevailing approach trains vision-language models on large-scale state--action trajectories via SFT. For GUI grounding, CogAgent~\cite{cheng2024cogagent} introduced spatial-aware visual understanding, SeeClick~\cite{cheng2024seeclick} provided explicit grounding supervision, and UGround~\cite{wang2025uground} achieved cross-platform universal grounding. For unified action prediction, ShowUI~\cite{lin2024showui} combined vision-language-action modeling, and UI-TARS~\cite{qin2025uitars} added reasoning-oriented fine-tuning. Mobile-domain extensions include Mobile-Agent-v3~\cite{wang2025mobileagentv3}, Aria-UI~\cite{huang2025ariaui}, and MAI-UI~\cite{huang2025maiui}, which target the unique challenges of Android interaction such as app diversity, dynamic layouts, and touch-based navigation. GUI-Libra~\cite{wang2025golibra} introduced RL-augmented supervision for grounding tasks.

\paragraph{Reinforcement learning for GUI agents.}
PPO~\cite{schulman2017ppo,ouyang2022instructgpt,yao2023react,wei2022cot} requires a value network; GRPO~\cite{shao2024deepseekmath} computes group-relative advantages instead. After DeepSeek-R1~\cite{guo2025deepseek} showed emergent reasoning from pure GRPO, the approach was extended to VLMs~\cite{huang2025visionr1,reinke2025reinforcedmllm,sun2025twostage}. In GUI, UI-R1~\cite{lu2025uir1} first applied rule-based GRPO without reasoning traces. Follow-ups explored multi-turn RL~\cite{zhang2025tars2,huang2025infigui}, experience replay~\cite{lu2025arpo}, curriculum~\cite{wang2025craft}, decoupled training~\cite{liu2025decoupled}, mobile-specific RL~\cite{li2025mobilegui,chen2025mobilerl}, self-evolutionary grounding~\cite{li2025segui}, efficient training~\cite{li2025efficient}, and alternative formulations~\cite{zhang2025uiagile,wang2025golibra}. Similar ideas appeared in web navigation~\cite{liu2025webagentr1} and environment-free training~\cite{shen2025simia}. While these methods develop genuine planning beyond imitation, all still require costly real-environment interaction.

\paragraph{World models for GUI.}
World models enable action-conditioned future prediction for planning~\cite{ha2018world,wei2025worldmodel,huang2025embodiedwm}. GUI approaches fall into three categories: \emph{text-based} (language/DOM prediction~\cite{chae2024guiworldmodel,gu2024llmwebworldmodel}), \emph{pixel-based} (diffusion-generated screenshots~\cite{luo2025vimo,chen2026mobiledreamer}), and \emph{code-based} (renderable HTML~\cite{zheng2026code2world}). UI-Simulator~\cite{zhang2025uisimulator} and SimuRA~\cite{liu2025simura} use world models for agent training simulation. All, however, operate at \emph{inference time} only~\cite{zheng2026code2world,luo2025vimo}. WM-R1 instead uses the world model as the \emph{training environment} itself.

%% file: sections/3_preliminaries.tex
\section{Preliminaries}
\label{sec:preliminaries}

\begin{figure*}[t]
\centering
\includegraphics[width=\linewidth]{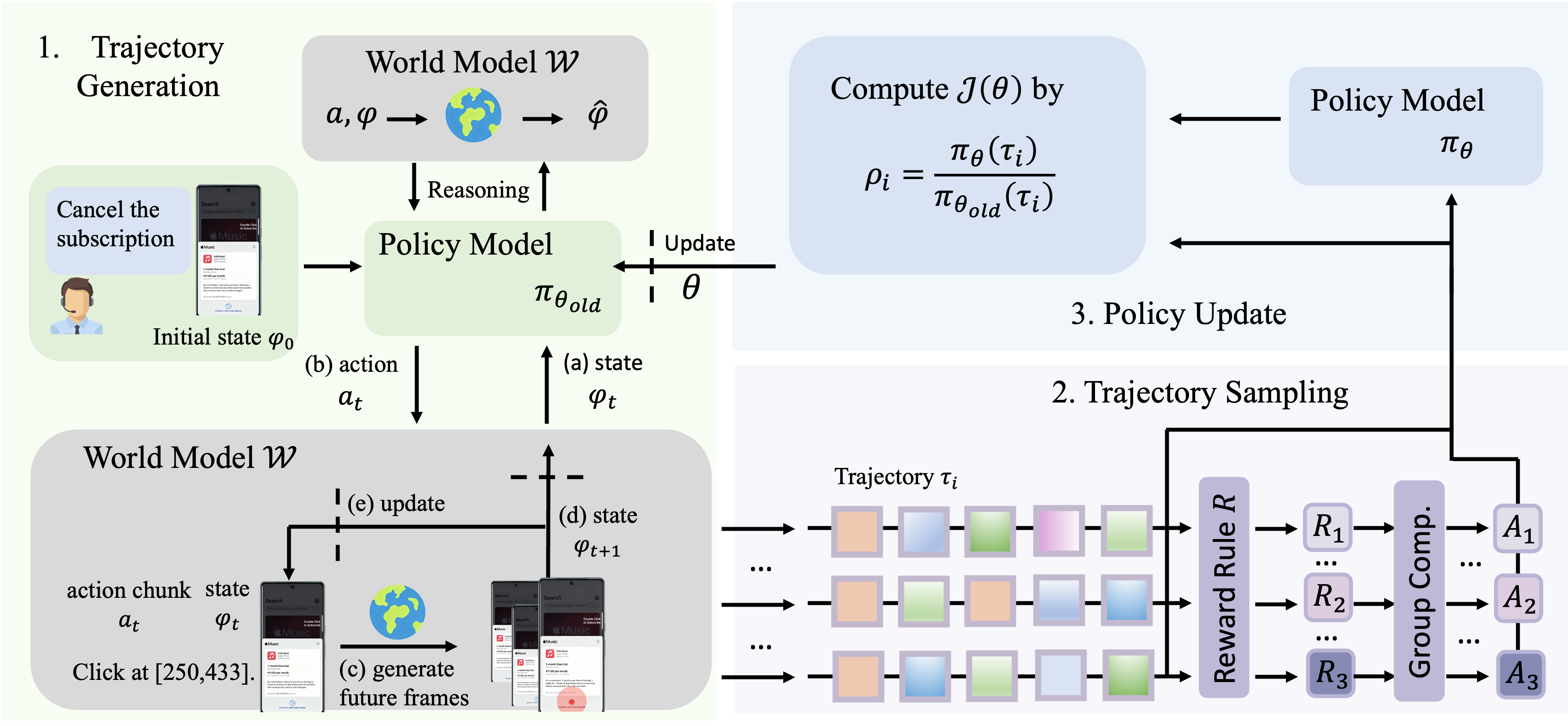}
\caption{WM-R1 training framework. For each task, the policy model generates multiple candidate trajectories with chain-of-thought reasoning in a world-model-simulated environment. At each reasoning step, the agent can invoke the world model to predict the consequence of a candidate action. The simulated next state becomes part of the trajectory. After all trajectories complete, a rule-based Reward Model computes scores, followed by GRPO advantage computation and policy update.}
\label{fig:framework}
\end{figure*}
At step $t$ in standard RL training, a GUI agent $\pi_\theta$ parameterized by $\theta$ observes a screenshot $\vphi_t$ and interaction history $\hist_{t-1} = (\vphi_0, \act_0, \ldots, \vphi_{t-1}, \act_{t-1})$, and produces an action $\act_t \in \gA$ (e.g., click, type, scroll). The environment transitions to a new state $\vphi_{t+1} \sim \gT(\cdot \mid \vphi_t, \act_t)$, yielding a sparse reward $\rew_t \in \{0, 1\}$. 
Training proceeds via Group Relative Policy Optimization (GRPO)~\cite{shao2024deepseekmath}, which eliminates the value network by computing advantages relative to a group of $G$ sampled trajectories. For each trajectory $\traj_i$ with reward $r_i$, the advantage is:
\begin{equation}
    A_i = \frac{r_i - \mu_r}{\sigma_r},
    \label{eq:grpo_advantage}
\end{equation}
where $\mu_r$ and $\sigma_r$ are the mean and standard deviation of rewards within the group. The policy update maximizes:
\begin{equation}
    \begin{aligned}[t]
    \mathcal{J}(\theta) &= \EE_{\traj \sim \pol_{\theta_{\text{old}}}} \Bigg[ \frac{1}{G} \sum_{i=1}^G \min\Big( \rho_i A_i, \\
    &\quad \text{clip}(\rho_i, 1-\epsilon, 1+\epsilon) A_i \Big) - \beta \cdot \KL\!\left(\pol_\theta \,\|\, \pol_{\text{ref}}\right) \Bigg],
    \end{aligned}
    \label{eq:grpo_objective}
\end{equation}
where $\rho_i = \pol_\theta(\traj_i) / \pol_{\theta_{\text{old}}}(\traj_i)$ is the importance sampling ratio and $\beta$ controls the KL divergence penalty from a reference policy $\pol_{\text{ref}}$.
WM-R1 replaces the real environment with world models $\wm$ making the simulator an integral component in the training loop. Concretely, this substitution enables two key capabilities beyond standard GRPO: (1) \emph{reasoning with simulated outcomes}: the agent invokes $\wm$ during its chain-of-thought to preview candidate actions, observe the predicted next state $\hat{\vphi}_{t+1}$, and refine its decision before committing to a final action $\act_t$; (2) \emph{fully simulated state transformation}: the entire trajectory $\tau_i$ is generated via state transformation $\hat{\vphi}_{t+1} \sim \wm(\cdot \mid \vphi_t, \act_t)$.

%% file: sections/4_method.tex
\section{Method: WM-R1}
\label{sec:method}

As illustrated in Figure~\ref{fig:framework}, WM-R1 trains a multimodal GUI agent through GRPO in a fully simulated environment powered by a world model. This substitution enables three things: (1) the agent learns to reason using world model predictions during its chain-of-thought (Section~\ref{subsec:reason}), (2) trajectory generation is massively parallelized, producing states without requiring real environments (Section~\ref{subsec:traj}), and (3) the reward design can focus on reasoning performance and action correctness (Section~\ref{subsec:reward}).

\subsection{Reasoning with the World Model}
\label{subsec:reason}
The core of WM-R1 is how the agent learns to invoke and reason with the world model during its chain-of-thought process. The policy model $\pol_\theta$ is prompted to produce responses in a structured format that interleaves reasoning tokens with actions. The system prompt is shown below:

\medskip
\begin{center}
\fbox{\begin{minipage}{0.93\linewidth}
\sffamily\small
\textbf{Prompt for Training and Inference}\par
\vspace{0.15em}
In this GUI screenshot, I want to perform the command \emph{instruction}. Please provide the action to perform and the coordinate where the cursor is moved to if click is performed. Output the thinking process in thinking tags and final answer in answer tags. Within the thinking process, you may invoke \textless call\_wm\textgreater to simulate the outcome of an action before committing. Please strictly follow the format.
\end{minipage}}
\end{center}
\medskip

The world model is integrated into the reasoning process through a \texttt{<call\_wm>} invocation. Unlike approaches that treat the world model as a separate conversational turn, WM-R1 embeds world model calls \emph{within} the agent's reasoning stream. At each reasoning step, the agent can decide to invoke the world model to simulate the consequence of a proposed action:
\begin{equation}
    \hat{\vphi}_{t+1} \sim \wm(\cdot \mid \hat{\vphi}_t, \act_t).
    \label{eq:wm_transition}
\end{equation}
The simulated screenshot $\hat{\vphi}_{t+1}$ is then injected back as a \emph{System (WM)} observation, and the agent continues reasoning from this point within the same thinking block.
This multi-turn reasoning pattern allows the agent to interleave thinking and world model calls within a single thinking block, then produce a final action once it is confident. 
% \begin{quote}
% \small
% \texttt{<\textbackslash think>}
% I need to click the search bar to enter the query. It appears on the left side of the top bar. Let me simulate clicking there.
% \texttt{<call\_wm>}(click(0.7, 0.3))
% \texttt{<\textbackslash think>}

% System (WM): [simulated screenshot $\hat{\vphi}_{t+1}$: nothing changed --- the click landed on an empty area to the right of the search bar]

% \texttt{<\textbackslash think>}
% The click did not hit the target. The coordinate was too far to the right. The search bar starts around x=0.4. Let me adjust.
% \texttt{<call\_wm>}(click(0.5, 0.3))
% \texttt{<\textbackslash think>}

% System (WM): [simulated screenshot $\hat{\vphi}_{t+2}$: search bar focused with cursor visible]

% \texttt{<\textbackslash think>}
% The search bar is now focused and the cursor is active. This is the correct action.
% \texttt{<\textbackslash /think>}

% \texttt{<action>}click(0.5, 0.3)\texttt{</action>}
% \end{quote}

\subsection{Trajectory Generation}
\label{subsec:traj}
Each training trajectory is generated through an iterative loop between the policy model and the world model. Given a task instruction $\instr$ and an initial screenshot $\vphi_0$, the trajectory generation proceeds as follows:

\begin{enumerate}
    \item Observe. The agent receives the current simulated screenshot $\hat{\vphi}_t$ along with the full interaction history $\hist_t = (\vphi_0, \act_0, \hat{\vphi}_1, \act_1, \ldots, \hat{\vphi}_t)$.

    \item Think. The policy model $\pol_\theta$ generates a reasoning block with optional \texttt{<call\_wm>} invocations:
    \begin{equation}
        \text{thought}_t \sim \pol_\theta(\hat{\vphi}_t, \instr, \hist_t).
        \label{eq:policy_think}
    \end{equation}

    \item Simulate \& Continue. If \texttt{<call\_wm>} is present, the world model generates the next simulated screenshot, which is fed back as a \textbf{System (WM)} observation. The agent continues reasoning from this point. If no \texttt{<call\_wm>} is present, the reasoning block ends and the agent produces a final action:
    \begin{equation}
        \act_t \sim \pol_\theta(\hat{\vphi}_t, \instr, \hist_t, \text{thought}_t).
    \end{equation}

    \item Repeat. Steps 1--3 continue until the agent produces a terminal action or the trajectory reaches the maximum length $T_{\max}$.
\end{enumerate}

The complete trajectory for task $i$ and rollout $j$ is:
\begin{equation}
    \traj_{i,j} = (\hat{\vphi}_0, \act_0, \hat{\vphi}_1, \act_1, \ldots, \hat{\vphi}_T, \act_T),
\end{equation}
The complete WM-R1 training loop is summarized in Algorithm~1 of the Supplementary Material.

\subsection{Reward Design}
\label{subsec:reward}
The reward function drives the GRPO optimization and determines what behaviors the agent learns. We adopt a composite reward design that combines a task success signal with a length penalty and a world model efficiency bonus:
\begin{equation}
    R = \alpha \cdot R_{\text{success}} + \beta \cdot R_{L} + \gamma \cdot R_{\text{WM}}.
    \label{eq:reward}
\end{equation}

\paragraph{Success reward $R_{\text{success}}$.}
The success reward evaluates whether the agent completes the task. We use LLM-as-a-judge to compare the final trajectory against the reference solution. The reward is binary:
\begin{equation}
    R_{\text{success}} =
    \begin{cases}
        1, & \text{task succeeds (exact match)} \\
        0, & \text{task fails}
    \end{cases}
\end{equation}
Exact match is determined by the LLM judge evaluating whether the sequence of actions and their effects are semantically equivalent to the reference trajectory.

\paragraph{Length reward $R_{L}$.}
The length reward encourages efficiency by penalizing unnecessarily long trajectories. Following DAST adopted in UI-R1~\cite{shen2025dast,lu2025uir1}, we define a length budget $L_{\text{budget}}$ as a mixture of the reference trajectory length $L_{\bar{r}}$ and the maximum allowed length $L_{\max}$:
\begin{equation}
    L_{\text{budget}} = p \cdot L_{\bar{r}} + (1 - p) \cdot L_{\max}, \qquad p = \frac{c}{N},
\end{equation}
where $c$ is the number of completed sub-tasks and $N$ is the total number of sub-tasks. The normalized deviation $\lambda = (L_i - L_{\text{budget}}) / L_{\text{budget}}$ is then used to compute the length reward:
\begin{equation}
    R_L =
    \begin{cases}
        \max(-0.5\lambda + 0.5,\; 0.1) & \text{if task succeeds} \\
        \min(0.9\lambda - 0.1,\; -0.1) & \text{if task fails}
    \end{cases}
\end{equation}

\paragraph{World model call count reward $R_{\text{WM}}$.}
The WM call reward regulates the agent's usage of world model simulation. Let $N_{\text{WM}}^{(i)}$ denote the number of \texttt{<call\_wm>} invocations in trajectory $i$. The group average is $\bar{N}_{\text{WM}} = \frac{1}{G}\sum_{j=1}^G N_{\text{WM}}^{(j)}$. The normalized deviation $\mu = (N_{\text{WM}}^{(i)} - \bar{N}_{\text{WM}}) / \bar{N}_{\text{WM}}$ is used to compute:
\begin{equation}
    R_{\text{WM}} =
    \begin{cases}
        \max(-0.5\mu + 0.5,\; 0.1) & \text{if task succeeds} \\
        \min(0.9\mu - 0.1,\; -0.1) & \text{if task fails}
    \end{cases}
\end{equation}

For each task, the $G$ trajectory rewards are used to compute GRPO advantages via Eq.~\eqref{eq:grpo_advantage}. The relative nature of GRPO advantages means that the agent learns to produce trajectories that are better than the group average, which naturally selects for more accurate and efficient reasoning strategies.
Through GRPO training, the agent learns \emph{when} and \emph{how many times} to invoke the world model: trajectories that use simulation effectively to refine their actions receive higher rewards, while unnecessary or poorly timed world model calls do not contribute positively. Over training, the agent internalizes a policy that balances reasoning depth with action efficiency.

%% file: sections/5_experiments.tex
\section{Experiments}
\label{sec:experiments}

\subsection{Experimental Setup}
\begin{table*}[t]
\centering
\resizebox{\textwidth}{!}{%
\setlength{\tabcolsep}{5pt}
\small
\begin{tabular}{lcccccccccc}
\toprule
& \multicolumn{2}{c}{\textbf{AndroidWorld}} & \multicolumn{2}{c}{\textbf{GUI-Odyssey}} & \multicolumn{2}{c}{\textbf{ScreenSpot-Pro}} & \multicolumn{2}{c}{\textbf{ScreenSpot-V2}} & \multicolumn{2}{c}{\textbf{AndroidControl}} \\
\textbf{Method} & \textbf{Full} & \textbf{Hard} & \textbf{Full} & \textbf{Hard} & \textbf{Icon} & \textbf{Text} & \textbf{Icon} & \textbf{Text} & \textbf{Type} & \textbf{Gnd} \\
\midrule
\multicolumn{11}{l}{\textit{3B scale}} \\
Qwen2.5-VL-3B (ZS) & 15.2 & 11.8 & 8.5 & 5.2 & 28.4 & 38.6 & 64.5 & 88.2 & 76.2 & 69.5 \\
Qwen2.5-VL-3B + CW & 20.6 & 16.3 & 14.2 & 9.8 & 31.2 & 41.5 & 67.8 & 89.1 & 78.5 & 72.3 \\
GUI-R1-3B & 17.4 & 13.5 & 10.2 & 6.8 & 35.6 & 44.3 & 76.8 & 90.5 & 82.3 & 75.6 \\
UI-TARS-3B & 19.8 & 15.6 & 13.4 & 8.9 & 39.2 & 48.5 & 80.4 & 91.8 & 84.7 & 77.8 \\
UI-R1-3B & 18.6 & 14.2 & 11.3 & 7.4 & 38.7 & 48.2 & 81.2 & 92.5 & 85.6 & 78.4 \\
\textbf{WM-R1-3B} & \textbf{29.5} & \textbf{23.1} & \textbf{22.8} & \textbf{16.4} & \textbf{43.5} & \textbf{52.8} & \textbf{83.6} & \textbf{93.8} & \textbf{89.2} & \textbf{82.4} \\
\midrule
\multicolumn{11}{l}{\textit{7B scale}} \\
Qwen2.5-VL-7B (ZS) & 22.4 & 17.6 & 12.3 & 7.8 & 35.2 & 44.8 & 72.3 & 91.4 & 79.8 & 74.1 \\
Qwen2.5-VL-7B + CW & 28.7 & 23.1 & 19.6 & 14.2 & 38.6 & 48.1 & 76.8 & 92.1 & 82.5 & 77.6 \\
GUI-R1-7B & 24.5 & 19.2 & 15.8 & 10.5 & 40.2 & 49.6 & 82.5 & 92.8 & 85.2 & 78.5 \\
UI-TARS-7B & 27.3 & 21.5 & 18.2 & 12.8 & 43.5 & 53.2 & 85.8 & 93.5 & 87.6 & 80.4 \\
Aguvis-7B & 26.8 & 20.9 & 17.5 & 12.1 & 44.2 & 54.5 & 86.5 & 94.2 & 88.2 & 81.5 \\
OS-Genesis-7B & 29.2 & 22.8 & 19.8 & 14.5 & 41.8 & 51.2 & 84.2 & 93.5 & 86.5 & 79.8 \\
UI-R1-7B & 30.8 & 24.2 & 21.5 & 15.3 & 42.8 & 52.4 & 84.5 & 93.6 & 87.4 & 81.2 \\
MobileGUI-RL-7B & 30.0 & 23.5 & 20.8 & 15.1 & 40.5 & 50.2 & 82.1 & 92.8 & 85.8 & 79.5 \\
OS-Atlas-7B & 33.5 & 27.2 & 24.8 & 19.2 & 45.8 & 55.6 & 88.2 & 94.8 & 90.5 & 84.2 \\
UI-TARS 1.5-7B & 34.2 & 28.1 & 25.6 & 20.1 & 46.2 & 56.8 & 89.5 & 95.1 & 91.2 & 85.5 \\
\textbf{WM-R1-7B} & \textbf{39.8} & \textbf{32.7} & \textbf{31.6} & \textbf{24.1} & \textbf{48.6} & \textbf{58.2} & \textbf{90.4} & \textbf{95.2} & \textbf{93.5} & \textbf{87.8} \\
\bottomrule
\end{tabular}}
\caption{Main results across all benchmarks. Bold: best per scale. AndroidWorld and GUI-Odyssey report success rate (\%); Hard subsets contain tasks requiring $\geq$5 ground-truth steps. ScreenSpot-Pro/V2 report mobile grounding accuracy (\%). AndroidControl reports action type prediction (Type) and coordinate grounding (Gnd) accuracy (\%). CW: Code2World inference-time augmentation.}
\label{tab:main_results}
\end{table*}
\paragraph{Datasets.}
We train the Qwen2.5-VL-3/7B model~\cite{bai2025qwen25vl} on a curated dataset combining AndroidCode~\cite{zheng2026code2world}, GUI-Odyssey~\cite{lu2024guiodyssey}, and GUI-R1~\cite{gui_r1_2025}. Tasks with zero-shot success rates above 80\% (trivially easy) or below 5\% (impossibly hard) are removed, and 2000 examples are randomly sampled, ensuring the training distribution focuses on intermediate-complexity tasks where world-model feedback is most informative.
\paragraph{Benchmarks.}
We evaluate on two categories of benchmarks. \emph{Long-horizon tasks}: AndroidWorld~\cite{rawles2024androidinthewild}, testing multi-step mobile device control with a maximum execution length of 15; and GUI-Odyssey~\cite{lu2024guiodyssey}, evaluating cross-app navigation on Android. For both AndroidWorld and GUI-Odyssey, we define a \emph{Hard} subset consisting of tasks that require $\geq$5 ground-truth steps to complete. \emph{Action prediction and grounding}: ScreenSpot-Pro~\cite{cheng2024screenspot} and ScreenSpot-V2~\cite{wu2024screenspotv2} evaluate GUI grounding accuracy on the mobile subset; AndroidControl~\cite{rawles2024androidinthewild} measures action type prediction and coordinate grounding.

\paragraph{Implementation details.}
We use the pretrained Code2World-8B model~\cite{zheng2026code2world}, frozen during training. It generates renderable HTML from the current screenshot and action, rendered to a $1080 \times 2400$ screenshot using a headless browser.
Full training configuration is provided in the Supplementary Material (Section~A).

\subsection{Main Results}

\paragraph{Long-horizon tasks.}
The results in Table~\ref{tab:main_results} demonstrate the advantages of world-model-simulated training for long-horizon mobile tasks.
WM-R1 establishes new state-of-the-art results at both model scales. WM-R1-3B (29.5) substantially outperforms all 3B-scale baselines including UI-R1-3B~\cite{lu2025uir1} (18.6) and GUI-R1-3B~\cite{gui_r1_2025} (17.4). At 7B scale, WM-R1-7B reaches 39.8 on AndroidWorld, outperforming UI-R1-7B~\cite{lu2025uir1} (30.8) by +9.0 points and UI-TARS 1.5-7B (34.2) by +5.6 points. On GUI-Odyssey, WM-R1-7B achieves 31.6, surpassing UI-R1-7B (21.5) by +10.1 points and OS-Atlas-7B (24.8) by +6.8 points.

WM-R1 achieves substantial gains over the inference-time augmentation baseline (Qwen2.5-VL + CW, i.e., using Code2World for propose-simulate-select at inference only) at both scales: +8.9 and +11.1 on AndroidWorld for 3B and 7B respectively. The improvements are particularly strong on the Hard subsets (+6.8 at 3B, +9.6 at 7B for AndroidWorld; +6.6 at 3B, +8.8 at 7B for GUI-Odyssey), indicating that world-model-simulated transitions provide especially strong signal for difficult, multi-step reasoning tasks that require multiple steps.
\begin{figure}[t]
\centering
\includegraphics[width=\linewidth]{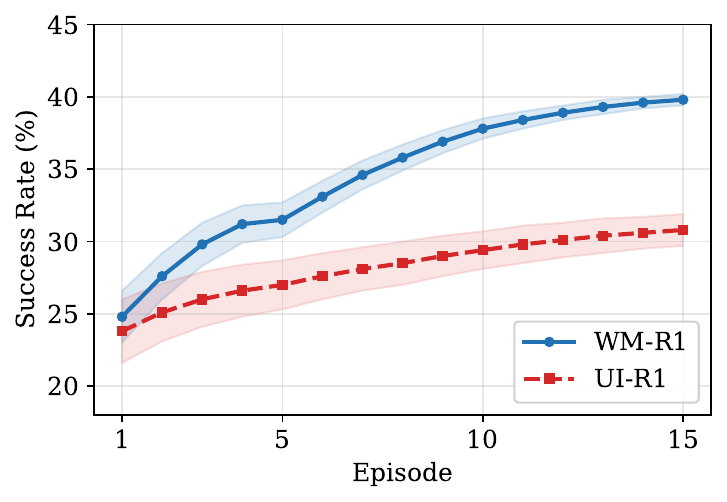}
\caption{Training curves. WM-R1 reaches 39.8\% success rate by episode 15, outperforming UI-R1 (30.8\%) by 9.0 points with lower variance.}
\label{fig:training_curve}
\end{figure}

A key question is whether WM-R1's improvements stem from memorizing training patterns or from learning generalizable reasoning strategies. To investigate this, we categorize our evaluation benchmarks into in-distribution (ID) and out-of-distribution (OOD) settings relative to the training data. AndroidControl and GUI-Odyssey serve as ID benchmarks while AndroidWorld and ScreenSpot-Pro/V2 are OOD.
WM-R1 achieves an average OOD improvement of +16.0 relative to ZS baseline, which is \emph{1.84$\times$} that of UI-R1 (+8.7). This demonstrates that world-model-based reasoning delivers stronger generalization to novel environments through two complementary mechanisms:
(i) World-model-based reasoning over memorization. Standard GRPO trains the agent to produce correct actions on the training distribution. When faced with unfamiliar UI elements or app combinations, the agent must rely on surface-level pattern matching, which degrades rapidly under distribution shift. WM-R1, by contrast, trains the agent to use the world model as an internal simulator: before committing to an action, the agent proposes a candidate, simulates its consequence via \texttt{<call\_wm>}, and revises if the predicted outcome does not match the intent. This ``think-before-act'' loop is a \emph{reasoning strategy}, not a memorized mapping. It transfers naturally to OOD situations because the agent can simulate and evaluate actions even on UI patterns it has never seen during training.
(ii) Diverse training distribution. The world model enhances the visual and structural diversity of the training distribution: by simulating the consequences of actions on diverse screenshots, the agent effectively trains on a broader and more novel distribution of state transitions than the raw dataset provides.
Together, these mechanisms explain why WM-R1 is expected to close the gap between ID and OOD performance more effectively than GRPO-only baselines: the agent learns \emph{how to reason with a world model} rather than \emph{what actions to take on specific UI patterns}.

\begin{figure}[t]
\centering
\includegraphics[width=\linewidth]{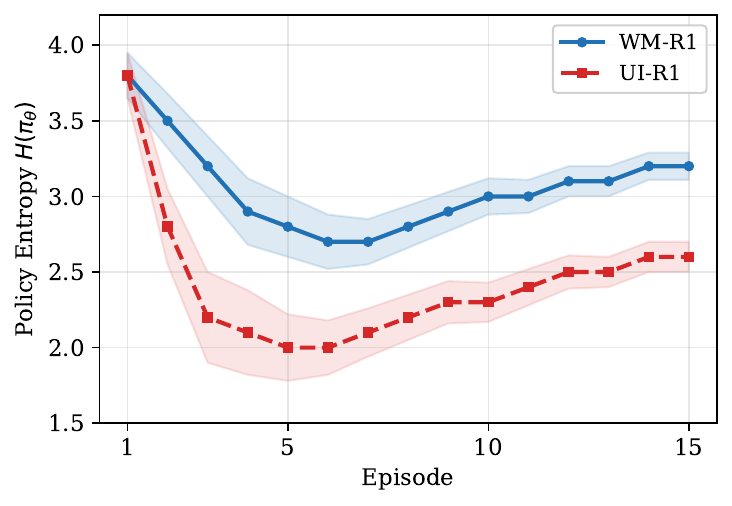}
\caption{Policy entropy. Both methods show a decrease-then-recover pattern, but WM-R1 maintains higher entropy throughout, indicating sustained exploration from the noise-free world model.}
\label{fig:entropy}
\end{figure}

Figure~\ref{fig:training_curve} shows the training dynamics over 15 episodes on AndroidWorld. Both methods start from the same zero-shot baseline (22.4\%), but WM-R1 exhibits a steeper learning curve, reaching 31.5\% by episode 5. By episode 15, WM-R1 reaches 39.8\%, a 9.0 point advantage over UI-R1. WM-R1 also shows lower variance across seeds, which we attribute to the fully simulated environment eliminating the stochasticity inherent in real GUI interactions. The policy entropy $H(\pi_\theta)$ (Figure~\ref{fig:entropy}) reveals the exploration mechanism behind this advantage: both methods exhibit an initial entropy drop followed by recovery, but WM-R1's entropy decreases much more gradually (3.8$\to$2.9 vs.\ UI-R1's 3.8$\to$2.1 in the first 3 episodes). This sustained exploration is enabled by the noise-free world model, which avoids the stochastic punishment of exploratory actions that causes real-environment agents to prematurely collapse. At convergence, WM-R1 retains higher entropy ($\sim$3.2 vs.\ $\sim$2.6).
\paragraph{Action prediction and grounding.}
WM-R1 delivers substantial improvements across all mobile grounding benchmarks. At 3B scale, WM-R1-3B improves icon grounding on ScreenSpot-Pro by +15.1 points (43.5 vs.~28.4) and text grounding by +14.2 points (52.8 vs.~38.6) over the zero-shot baseline, while surpassing UI-R1-3B~\cite{lu2025uir1} across all metrics. On AndroidControl, WM-R1-3B achieves 89.2 action type accuracy and 82.4 coordinate grounding (avg 85.8), exceeding UI-R1-3B (85.6 / 78.4, avg 82.0) by +3.8 points. At 7B scale, WM-R1-7B further extends these gains: ScreenSpot-Pro icon reaches 48.6 (+13.4 over ZS), ScreenSpot-V2 icon reaches 90.4 (+18.1 over ZS), and AndroidControl grounding reaches 87.8 (+13.7 over ZS). The grounding improvement is driven by the world model's dense visual feedback: by simulating the consequence of each candidate action, the agent learns to associate visual features with spatial locations more precisely.

\subsection{Ablation Studies}

To understand the contribution of each component, we conduct an ablation study on AndroidWorld (7B scale). We evaluate three variants of WM-R1, each removing a single design choice from the full system.

\begin{table}[t]
\centering
\setlength{\tabcolsep}{4pt}
\setlength{\extrarowheight}{4pt}
\small
\begin{tabular}{lccc}
\toprule
\textbf{Config} & \textbf{AW} & \textbf{AW-H} & $\Delta$ \\
\midrule
WM-R1 (full) & \textbf{39.8} & \textbf{32.7} & --- \\
$-$ CoT & 34.6 & 27.8 & $-$5.2 / $-$4.9 \\
$-$ $R_L$ & 36.5 & 30.1 & $-$3.3 / $-$2.6 \\
$-$ $R_{\text{WM}}$ & 37.2 & 30.8 & $-$2.6 / $-$1.9 \\
\bottomrule
\end{tabular}
\caption{Ablation study on AndroidWorld (7B). Chain-of-thought reasoning is the most critical component; removing it causes the largest degradation ($-$5.2 on AW, $-$4.9 on AW-H). $R_L$ and $R_{\text{WM}}$ contribute smaller but complementary gains.}
\label{tab:ablation}
\end{table}

Table~\ref{tab:ablation} reveals the relative importance of WM-R1's components. Chain-of-thought reasoning is the most impactful, with a 5.2 point degradation on AndroidWorld and 4.9 on the Hard subset when removed. Without intermediate reasoning, the model produces actions directly from the observation, bypassing the reflective loop where the world model's simulated feedback can be used to evaluate and revise candidate actions before execution. The length reward $R_L$ contributes a 3.3 point gain: without it, the agent learns to solve tasks correctly but with redundant steps that accumulate error over longer trajectories. The WM call reward $R_{\text{WM}}$ has the smallest individual impact ($-$2.6 on the full set, $-$1.9 on Hard), but its role is stabilizing: without regulation, some training trajectories converge to excessive world model calls that waste inference budget, while others underuse the simulator entirely. Together, the three components address distinct failure modes, and the full system achieves a 5.2 point advantage over the strongest single-component variant.

\subsection{World Model Utilization Dynamics}
\label{subsec:wm_dynamics}

A natural question is how the agent's use of the world model evolves during training. We track two quantities across training episodes: (1) the average number of \texttt{<call\_wm>} invocations per trajectory, and (2) the average episode reward.

\begin{figure}[t]
\centering
\includegraphics[width=\linewidth]{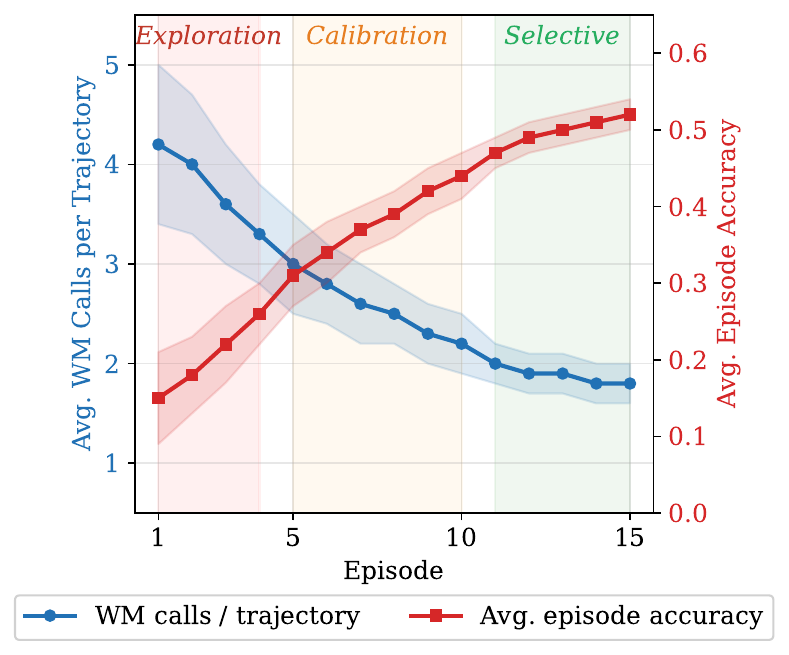}
\caption{WM utilization dynamics. WM calls per trajectory (blue, left axis) decrease from $\sim$4.2 to $\sim$1.8 while average episode reward (red, right axis) increases from 0.15 to 0.52. Three phases emerge: exploration (ep.\ 1--4), calibration (ep.\ 5--10), and selective utilization (ep.\ 11--15).}
\label{fig:wm_dynamics}
\end{figure}

As shown in Figure~\ref{fig:wm_dynamics}, a clear inverse relationship emerges: as training progresses, the agent makes \emph{fewer} world model calls while achieving \emph{higher} rewards. We identify three phases of world model mastery:

\textbf{Exploration phase} (episodes 1--4): The agent invokes the world model frequently ($\sim$4 calls per trajectory), often redundantly simulating obvious actions. Many of these calls produce predictions that the agent could have anticipated without simulation, reflecting an undifferentiated reliance on the world model.

\textbf{Calibration phase} (episodes 5--10): The agent begins to distinguish between actions that benefit from simulation (e.g., clicking on ambiguous UI elements, navigating to unfamiliar screens) and actions that do not (e.g., typing text, pressing the home button). World model calls drop to $\sim$2.5 per trajectory as the agent learns to allocate simulation budget strategically.

\textbf{Selective utilization phase} (episodes 11--15): The agent invokes the world model primarily at critical decision points---when it is uncertain about the consequence of an action or when a mistake would derail the remaining trajectory. Calls stabilize at $\sim$1.8 per trajectory, indicating that the agent has internalized a policy of \emph{targeted simulation}: using the world model where it provides the most value while relying on its own learned policy for routine actions.

This dynamic is consistent with the role of $R_{\text{WM}}$ in our composite reward design: by penalizing both over-use and under-use relative to the group average, the reward guides the agent toward an efficient equilibrium.
% The decreasing call count alongside increasing reward confirms that fewer, better-targeted simulations produce stronger outcomes---the agent learns not just to use the world model, but to use it \emph{wisely}.

\subsection{Case Study: World-Model-Guided Self-Correction}

To illustrate how WM-R1 agents leverage the world model during inference, we present a representative reasoning trace requiring the agent to click a search bar and enter a query.

\begin{figure}[ht]
\centering
\fbox{\begin{minipage}{0.92\linewidth}
\small\texttt{%
<think> \\
I need to click the search bar to enter the query. It appears on the left side of the top bar. Let me simulate clicking there. \\
\textless call\_wm\textgreater(click(0.7, 0.3)) \\
</think> \\[4pt]
System (WM): [simulated screenshot $\hat{\vphi}_{t+1}$: nothing changed --- the click landed on an empty area to the right of the search bar] \\[4pt]
<think> \\
The click did not hit the target. The coordinate was too far to the right. The search bar starts around x=0.4. Let me adjust. \\
\textless call\_wm\textgreater(click(0.5, 0.3)) \\
</think> \\[4pt]
System (WM): [simulated screenshot $\hat{\vphi}_{t+2}$: search bar focused with cursor visible] \\[4pt]
<think> \\
The search bar is now focused and the cursor is active. This is the correct action. \\
</think> \\[4pt]
Action: click(0.5, 0.3)
}
\end{minipage}}
% \caption{A representative reasoning trace from a WM-R1 agent. The agent first attempts a click at $(0.7, 0.3)$, simulates the outcome via \texttt{<call\_wm>}, observes that the click missed the target, revises the coordinate to $(0.5, 0.3)$, confirms the corrected action via a second world model call, and commits to the final action.}
\label{fig:case_study}
\end{figure}

The agent demonstrates a three-stage reasoning process: (1) Propose. It identifies the search bar and proposes an initial click coordinate; (2) Simulate and Evaluate. It invokes the world model to preview the consequence, and recognizes that the click landed on an empty area; (3) Revise and Confirm. It adjusts the coordinate, simulates again, and only commits once the predicted state matches the intent. This self-correction loop is impossible for GRPO-only baselines, which must execute actions directly and suffer irreversible consequences from initial mistakes. The world model effectively serves as a ``scratchpad'' that allows the agent to make and learn from mistakes before acting in the environment.

%% file: sections/6_conclusion.tex
\section{Conclusion}
\label{sec:conclusion}

We presented WM-R1, a reinforcement learning framework that trains GUI agents entirely within a world-model-simulated environment. Rather than using the world model only at inference time as a post-hoc proposal-evaluation tool, WM-R1 uses world models as the source of all state transitions during GRPO training. This design enables massively parallel, zero-cost training while allowing the agent's chain-of-thought reasoning to naturally incorporate simulated visual feedback at every step. Our experiments on mobile benchmarks demonstrate that WM-R1 improves over zero-shot and GRPO baselines, confirming that world-model-simulated training provides a practical and effective path toward generalizable mobile GUI reasoning agents.

%% file: sections/A_appendix.tex
\section{Appendix}
\subsection{Additional Training Details}
\label{app:training_details}

\paragraph{Hyperparameters.}
Table~\ref{tab:hyperparams} lists the full hyperparameter configuration used in our experiments.

\begin{table}[ht]
\centering
\begin{tabular}{lc}
\toprule
\textbf{Hyperparameter} & \textbf{Value} \\
\midrule
Base model & Qwen2.5-VL-3B / 7B \\
World model & Code2World-8B (frozen) \\
GPUs per node & 1 $\times$ NVIDIA H200 (141GB) \\
Number of nodes & 1 \\
World model instances & 1 \\
Rollout batch size & 2 \\
Group size ($G$) & 8 \\
Max prompt length & 64,000 \\
Max response length & 8,192 \\
Max pixels & 2,116,800 \\
Learning rate & $1 \times 10^{-6}$ \\
Learning rate schedule & Cosine decay \\
Warmup ratio & 0.05 \\
Gradient clip norm & 1.0 \\
PPO epochs & 1 \\
Clip ratio (low) & 0.2 \\
Clip ratio (high) & 0.3 \\
Max trajectory length & 15 \\
Total episodes & 15 \\
KL penalty $\beta$ & 0.01 \\
Reward: $\alpha$ (success) & 1.0 \\
Reward: $\beta$ (length) & 0.5 \\
Reward: $\gamma$ (WM call) & 0.1 \\
GPU memory utilization & 0.6 \\
temperature & 1.0 \\
\bottomrule
\end{tabular}
\caption{WM-R1 training hyperparameters.}
\label{tab:hyperparams}
\end{table}

\paragraph{Distributed training setup.}
We use Ray for distributed coordination. Each world model instance runs on a separate Ray actor. The actor-rollout worker group uses FSDP for memory-efficient training, with vLLM handling sequence generation during the rollout phase. The vLLM engine is put into sleep mode between generation phases to reduce peak memory usage.

\paragraph{World model configuration.}
We use the pretrained Code2World-8B model~\cite{zheng2026code2world}, frozen during training. It generates renderable HTML from the current screenshot and action, rendered to a $1080 \times 2400$ screenshot using a headless browser.

\paragraph{Data curation.}
We curate training data by combining three Android GUI datasets: AndroidCode~\cite{zheng2026code2world}, GUI-Odyssey~\cite{lu2024guiodyssey}, and GUI-R1~\cite{gui_r1_2025}. All three datasets are publicly available. We filter by task difficulty: tasks with zero-shot success rates above 80\% (trivially easy) or below 5\% (impossibly hard) are removed. From the filtered pool, we randomly sample 2000 examples, ensuring the training distribution focuses on intermediate-complexity tasks where world-model feedback is most informative.

\paragraph{Reward function details.}
Our reward function combines three components as described in the main paper:
\begin{itemize}
    \item \textbf{Success reward} $R_{\text{success}}$ (+1 if the task is completed as judged by an LLM comparing the final trajectory against the reference solution; 0 otherwise), weighted by $\alpha = 1.0$.
    \item \textbf{Length reward} $R_{L}$ encourages efficient trajectories by penalizing unnecessarily long ones, following the DAST approach~\cite{shen2025dast,lu2025uir1}, weighted by $\beta = 0.5$. The length budget $L_{\text{budget}}$ is a mixture of the reference trajectory length and the maximum allowed length, with the reward computed based on the normalized deviation from this budget.
    \item \textbf{WM call reward} $R_{\text{WM}}$ regulates the agent's usage of world model simulation by rewarding trajectories that use the group-average number of \texttt{<call\_wm>} invocations, weighted by $\gamma = 0.1$. Trajectories that overuse or underuse the world model relative to the group receive lower rewards.
\end{itemize}
The composite reward is $R = \alpha \cdot R_{\text{success}} + \beta \cdot R_{L} + \gamma \cdot R_{\text{WM}}$. GRPO advantages are computed group-relatively from these composite rewards.

\paragraph{Evaluator model and prompt design.}
The success reward $R_{\text{success}}$ is computed by an LLM-based evaluator. We use GPT-4o (version \texttt{gpt-4o-2024-08-06}) as the judge model with temperature 0 to maximize determinism. The evaluator receives a structured prompt containing four fields: (1) the original task instruction $I$, (2) the reference solution trajectory (ground-truth action sequence), (3) the agent's executed trajectory (action sequence with reasoning traces), and (4) the final simulated screenshot $\hat{\vphi}_T$. The judge is instructed to output a structured JSON response with two fields: a binary \texttt{success} flag (1 if the task objective is achieved, 0 otherwise) and a \texttt{reason} field providing a brief justification.

\paragraph{Calibration procedure.}
We calibrate the LLM evaluator on a held-out set of 200 human-annotated trajectories covering diverse task types (single-app operations, cross-app navigation, form filling, search tasks). For each trajectory, three independent annotators provide binary success/failure judgments, achieving a Fleiss' $\kappa$ of 0.89 (near-perfect agreement). We use majority vote as ground truth. The LLM evaluator achieves 93.5\% agreement with this human consensus on the calibration set. To further reduce variance, we average two independent LLM evaluations per trajectory; the inter-evaluation agreement rate is 96.2\%, and disagreements are resolved by majority vote with the reference trajectory.

\paragraph{Alignment with benchmark metrics.}
For AndroidWorld, we adopt the benchmark's official evaluation protocol, which combines programmatic state checks (e.g., verifying specific UI element states) with LLM-based judgment for tasks lacking deterministic success criteria. On ScreenSpot-Pro, ScreenSpot-V2, and AndroidControl, success is determined by exact coordinate or action type matching against ground truth, requiring no LLM evaluation. To validate consistency, we compare the LLM judge against programmatic string/attribute matching on a subset of 100 AndroidWorld tasks with deterministic success criteria: the agreement rate is 97.0\%, with disagreements concentrated on borderline cases where the agent partially completes the task. The same evaluator configuration is applied uniformly to all methods, ensuring that comparative results are not confounded by evaluation differences.

\subsection{WM-R1 Training Algorithm}
\label{app:algorithm}

Algorithm~\ref{alg:wmr1} presents the complete WM-R1 training loop.

\begin{algorithm}[ht]
\caption{WM-R1 Training Loop}
\label{alg:wmr1}
\begin{algorithmic}[1]
\REQUIRE Policy $\pol_\theta$, world model $\wm$, episodes $E$, group size $G$, batch size $B$
\FOR{episode $e = 1$ to $E$}
    \FOR{each batch of task instructions $\{I^{(i)}\}_{i=1}^B$}
        \STATE Initialize simulated states $\hat{\vphi}_0^{(i,j)} \gets \vphi_0^{(i)}$ for each task $i$ and rollout $j$
        \FOR{step $t = 1$ to $T_{\max}$}
            \FOR{each rollout $j \in \{1,\ldots,G\}$}
                \STATE Generate initial thought from $\pol_\theta(\hat{\vphi}_t^{(i,j)}, I^{(i)}, \hist_t^{(i,j)})$
                \WHILE{thought contains \texttt{<call\_wm>}}
                    \STATE Extract action $\act$ from \texttt{<call\_wm>} tag
                    \STATE $\hat{\vphi}_{t+1}^{(i,j)} \gets \wm(\hat{\vphi}_t^{(i,j)}, \act)$
                    \STATE Append System (WM) observation with $\hat{\vphi}_{t+1}^{(i,j)}$
                    \STATE Continue thought from $\pol_\theta(\hat{\vphi}_{t+1}^{(i,j)}, I^{(i)}, \hist_{t+1}^{(i,j)})$
                \ENDWHILE
                \IF{thought ends with \texttt{<action>}}
                    \STATE Extract final action $\act_t^{(i,j)}$
                \ELSE
                    \STATE Mark trajectory $j$ as complete
                \ENDIF
            \ENDFOR
            \IF{all rollouts complete}
                \STATE \textbf{break}
            \ENDIF
        \ENDFOR
        \STATE Compute composite rewards for all $G$ trajectories (main paper, Eq.~1)
        \STATE Compute GRPO advantages and update $\pol_\theta$ (main paper, Eqs.~2--3)
    \ENDFOR
\ENDFOR
\end{algorithmic}
\end{algorithm}

\subsection{Parameter Sensitivity Analysis}
\label{app:sensitivity}

We evaluate the sensitivity of WM-R1 to key hyperparameters on the AndroidWorld validation set using Qwen2.5-VL-7B. Table~\ref{tab:sensitivity} reports results when varying each parameter individually while holding others at the default configuration.

\begin{table}[ht]
\centering
\setlength{\tabcolsep}{4pt}
\small
\begin{tabular}{lcc}
\toprule
\textbf{Parameter} & \textbf{Value} & \textbf{AndroidWorld} \\
\midrule
\multirow{3}{*}{Reward $\alpha$ (success)} & 0.5 & 37.2 \\
& \textbf{1.0 (default)} & \textbf{39.8} \\
& 2.0 & 38.5 \\
\midrule
\multirow{3}{*}{Reward $\beta$ (length)} & 0.1 & 36.8 \\
& 0.3 & 38.1 \\
& \textbf{0.5 (default)} & \textbf{39.8} \\
\midrule
\multirow{3}{*}{Reward $\gamma$ (WM call)} & \textbf{0.1 (default)} & \textbf{39.8} \\
& 0.2 & 38.9 \\
& 0.4 & 37.6 \\
\midrule
\multirow{4}{*}{Group size $G$} & 4 & 37.5 \\
& \textbf{8 (default)} & \textbf{39.8} \\
& 16 & 39.2 \\
& 32 & 38.8 \\
\midrule
\multirow{3}{*}{KL penalty} & 0.005 & 36.5 \\
& \textbf{0.01 (default)} & \textbf{39.8} \\
& 0.05 & 37.8 \\
\midrule
\multirow{3}{*}{Max trajectory length} & 10 & 37.2 \\
& \textbf{15 (default)} & \textbf{39.8} \\
& 20 & 40.1 \\
\midrule
\multirow{3}{*}{Learning rate} & $5 \times 10^{-7}$ & 37.8 \\
& \textbf{$1 \times 10^{-6}$ (default)} & \textbf{39.8} \\
& $5 \times 10^{-6}$ & 35.8 \\
\bottomrule
\end{tabular}
\caption{Parameter sensitivity analysis on AndroidWorld (Qwen2.5-VL-7B). The \textbf{default} configuration is highlighted in each row.}
\label{tab:sensitivity}
\end{table}

WM-R1 shows moderate sensitivity to hyperparameter choices, with no single parameter causing catastrophic degradation:

\noindent\textbf{Reward coefficients.} The success reward weight $\alpha$ is the most influential of the three: setting it too low (0.5) dilutes the primary training signal, while setting it too high (2.0) makes the length and WM call rewards negligible. The length reward $\beta$ shows a clear optimum at 0.5 --- too low (0.1) allows bloated trajectories, too high over-penalizes necessary multi-step reasoning. The WM call reward $\gamma$ similarly peaks at 0.1, confirming that moderate regularization is optimal: the agent should use the world model as a tool, not as a crutch.

\noindent\textbf{Group size.} Performance peaks at $G=8$, consistent with the original GRPO findings. Smaller groups (4) provide insufficient variance for advantage estimation, while larger groups (16, 32) dilute the group-relative signal and increase memory overhead without corresponding gains.

\noindent\textbf{KL penalty.} The KL penalty is critical for training stability. Too small (0.005) allows the policy to drift too far from the reference, producing malformed outputs. Too large (0.05) constrains the policy too tightly, limiting the exploration needed to discover better reasoning strategies.

\noindent\textbf{Max trajectory length.} Increasing from 10 to 15 yields a notable gain, as longer trajectories allow the agent to complete more complex multi-step tasks. Extending to 20 provides minimal additional benefit, suggesting that 15 steps captures the majority of useful reasoning depth for AndroidWorld tasks.

\noindent\textbf{Learning rate.} The optimal learning rate is $1 \times 10^{-6}$. A higher rate ($5 \times 10^{-6}$) causes training instability and reward oscillation, while a lower rate ($5 \times 10^{-7}$) slows convergence, leaving the policy under-trained after 15 episodes.

Overall, the sensitivity analysis demonstrates that WM-R1 is robust to reasonable hyperparameter variation, with the default configuration representing a stable optimum across all dimensions.

\subsection{Computational Cost Analysis}
\label{app:cost}

Table~\ref{tab:cost} reports the computational cost of training and inference for WM-R1 compared with representative baselines.

\begin{table}[ht]
\centering
\setlength{\tabcolsep}{4pt}
\small
\begin{tabular}{lcccc}
\toprule
\textbf{Method} & \textbf{GPU} & \makecell{\textbf{Train}\\\textbf{hours}} & \makecell{\textbf{Env}\\\textbf{required}} & \makecell{\textbf{Inference}\\\textbf{+overhead}} \\
\midrule
UI-R1-7B (no env) & 1$\times$H200 & 7.8 & No & --- \\
MobileGUI-RL-7B & 1$\times$H200 & 36.5 & Yes & --- \\
WM-R1-3B & 1$\times$H200 & 8.4 & No & +1.6s/traj \\
WM-R1-7B & 1$\times$H200 & 11.2 & No & +2.3s/traj \\
\bottomrule
\end{tabular}
\caption{Computational cost comparison. ``GPU'' denotes hardware used; ``Train hours'' is wall-clock training time for the full 15-episode curriculum on the 2000-task dataset; ``Env required'' indicates whether a live Android environment is needed during training; ``Inference +overhead'' reports the additional latency per trajectory from world model calls at inference time relative to a standard generate-and-act baseline. UI-R1 (no env) refers to the offline GRPO variant that trains on pre-collected trajectories without real-environment interaction during training.}
\label{tab:cost}
\end{table}

\paragraph{Training cost breakdown.}
WM-R1-7B completes 15 training episodes on the 2000-task dataset in 11.2 wall-clock hours on a single NVIDIA H200 (141GB). The cost decomposes into: (i) world model simulation ($\sim$45\% of total time): Code2World-8B generates renderable HTML from each (screenshot, action) pair, followed by headless-browser rendering at $1080 \times 2400$ resolution ($\sim$1.1s per frame); (ii) policy generation ($\sim$35\%): the agent model generates reasoning traces and action predictions via vLLM with the sleep-mode optimization; (iii) training update ($\sim$20\%): FSDP-based forward/backward pass and parameter update.

\paragraph{Comparison with UI-R1 (no env).}
UI-R1-7B~\cite{lu2025uir1}, which uses offline GRPO on pre-collected trajectories without real-environment interaction during training, completes training in 7.8 GPU-hours on the same hardware and dataset. The additional 3.4 hours (1.44$\times$) for WM-R1 comes from world model simulation during rollouts. In return, WM-R1 achieves +9.0 points on AndroidWorld (39.8 vs.\ 30.8), representing a favorable cost-performance tradeoff. Notably, online RL methods such as MobileGUI-RL~\cite{li2025mobilegui} require a live Android simulator environment with real device resets, consuming $\sim$36.5 GPU-hours on the same hardware and requiring substantial infrastructure for environment management. WM-R1 eliminates all real-environment dependencies while running on a single GPU.

\paragraph{Inference overhead.}
At inference time, WM-R1 agents invoke the world model an average of 1.8 times per trajectory, adding approximately 2.3 seconds of latency per trajectory for the 7B model (1.6 seconds for 3B). This overhead is modest relative to typical task completion times (30--90 seconds per trajectory) and is offset by the substantial accuracy improvements demonstrated in the main results (Table~1 of the main paper).

\subsection{Generalization to Other Base Models (Qwen3-VL)}
\label{app:qwen3vl}

To verify that WM-R1's benefits are not specific to the Qwen2.5-VL architecture, we conduct additional experiments using the Qwen3-VL model family~\cite{bai2025qwen3vl} at the 4B and 8B scales.

\paragraph{Experimental setup.}
We apply the identical WM-R1 training configuration (Table~\ref{tab:hyperparams}) to Qwen3-VL-4B and Qwen3-VL-8B, using the same 2000-task curated dataset, world model (Code2World-8B), and reward weights. All other hyperparameters remain unchanged. Evaluation follows the same protocol as the main experiments.

\begin{table}[ht]
\centering
\setlength{\tabcolsep}{4pt}
\small
\begin{tabular}{lcccc}
\toprule
\textbf{Method} & \textbf{AW} & \textbf{AW-H} & \textbf{GO} & \textbf{GO-H} \\
\midrule
\multicolumn{5}{l}{\textit{4B scale}} \\
Qwen3-VL-4B (ZS) & 24.8 & 19.5 & 14.2 & 9.1 \\
Qwen3-VL-4B + CW & 31.2 & 25.4 & 21.5 & 15.8 \\
UI-R1-4B & 26.5 & 21.2 & 17.8 & 12.4 \\
\textbf{WM-R1-4B} & \textbf{38.2} & \textbf{31.5} & \textbf{30.2} & \textbf{23.4} \\
\midrule
\multicolumn{5}{l}{\textit{8B scale}} \\
Qwen3-VL-8B (ZS) & 28.5 & 22.8 & 17.5 & 11.6 \\
Qwen3-VL-8B + CW & 34.8 & 28.6 & 24.2 & 18.3 \\
UI-R1-8B & 35.2 & 28.5 & 25.8 & 19.6 \\
\textbf{WM-R1-8B} & \textbf{44.5} & \textbf{37.2} & \textbf{35.8} & \textbf{28.5} \\
\bottomrule
\end{tabular}
\caption{WM-R1 results with Qwen3-VL base models on long-horizon benchmarks. AW: AndroidWorld, GO: GUI-Odyssey, H: Hard subset. Bold: best per scale.}
\label{tab:qwen3vl}
\end{table}

\paragraph{Results.}
Table~\ref{tab:qwen3vl} shows that WM-R1 delivers consistent improvements across both Qwen3-VL scales. WM-R1-8B achieves 44.5 on AndroidWorld and 35.8 on GUI-Odyssey, surpassing UI-R1-8B (35.2 / 25.8) by +9.3 and +10.0 points respectively. At the 4B scale, WM-R1-4B reaches 38.2 on AndroidWorld, outperforming UI-R1-4B (26.5) by +11.7 points. The relative improvement pattern mirrors the Qwen2.5-VL results: gains are largest on Hard subsets and long-horizon tasks, confirming that world-model-simulated training is architecture-agnostic and benefits from stronger base models.

\paragraph{Choice of Qwen2.5-VL for main experiments.}
We use Qwen2.5-VL as the base model in the main experiments to ensure direct comparability with the majority of recent GUI agent work: UI-R1~\cite{lu2025uir1}, MobileGUI-RL~\cite{li2025mobilegui}, GUI-R1~\cite{gui_r1_2025}, and numerous other baselines in the main results table all adopt Qwen2.5-VL as their standard backbone. Using the same base model isolates the contribution of the training framework from differences in visual-language pretraining. The Qwen3-VL results above demonstrate that WM-R1 generalizes to newer architectures and in fact benefits from their improved capabilities, suggesting that future work can achieve further gains by combining WM-R1 with advancing foundation models.

\subsection{Limitations}

Despite its strengths, WM-R1 has several limitations:

\noindent\textbf{Dependence on world model fidelity.} The training signal is entirely derived from simulated state transitions produced by Code2World. While the world model achieves high visual fidelity for common UI transitions (button clicks, page navigation, form filling), it may struggle with highly dynamic or interactive elements (animations, real-time data updates, third-party widgets). Inaccurate world model predictions can mislead the agent during training, potentially causing it to learn suboptimal policies that do not generalize to real environments.

\noindent\textbf{Limited to discrete action spaces.} The current formulation assumes a discrete set of admissible actions at each step (click coordinates, typed text, scroll directions). Extending WM-R1 to continuous control settings (e.g., free-form cursor movement, drag-and-drop gestures) would require modifications to both the world model and the reward design.

\subsection{Broader Impact}

WM-R1 enables more efficient training of GUI agents, reducing the need for real-environment interaction and making it more feasible to train capable agents on commodity hardware, potentially democratizing access to agent-building tools for researchers with limited computational resources. The massively parallel simulation approach substantially reduces energy consumption compared to real-environment RL, as computation is concentrated on simulated transitions rather than costly environment step executions.

The world-model-simulated training paradigm also improves the interpretability of agent decision-making: the chain-of-thought reasoning stream explicitly records when and how the agent invokes the world model to evaluate candidate actions, providing a transparent signal that explains the agent's deliberation process. Beyond efficiency gains, the improved grounding and reasoning accuracy demonstrated on AndroidWorld and other mobile benchmarks suggests potential for practical applications in accessibility automation, repetitive workflow automation, and AI-assisted software testing. The deterministic nature of simulated transitions enhances reproducibility compared to methods relying on real-environment interaction, where non-deterministic state changes can make results difficult to replicate.